\documentclass[runningheads]{llncs}

\usepackage{eccv}

\usepackage{eccvabbrv}

\usepackage{graphicx}
\usepackage{booktabs}

\usepackage[accsupp]{axessibility}  

\usepackage[pagebackref,breaklinks,colorlinks,allcolors=eccvblue]{hyperref}

\usepackage{orcidlink}

\newcommand{\ourmethod}{\textit{AlbumFill}}

\begin{document}

\title{\ourmethod: Album-Guided Reasoning and Retrieval for Personalized Image Completion}

\titlerunning{\textit{AlbumFill}}

\author{
    Yu-Ju Tsai$^{1}$ \quad 
    Brian Price$^{2}$ \quad 
    Qing Liu$^{2}$ \quad 
    Luis Figueroa$^{2}$ \quad \\
    Daniil Pakhomov$^{2}$ \quad
    Zhihong Ding$^{2}$ \quad
    Scott Cohen$^{2}$ \quad
    Ming-Hsuan Yang$^{1}$ 
}

\authorrunning{Y.-J.~Tsai et al.}

\institute{University of California, Merced \and
Adobe Research}

\maketitle

\begin{abstract}
Personalized image completion aims to restore occluded regions in personal photos while preserving identity and appearance. 
Existing methods either rely on generic inpainting models that often fail to maintain identity consistency, or assume that suitable reference images are explicitly provided. 
In practice, suitable references are often not explicitly provided, requiring the system to search for identity-consistent images within personal photo collections.
We present \textit{AlbumFill}, a training-free framework that retrieves identity-consistent references from personal albums for personalized completion. 
Given an occluded image and a personal album, a vision-language model infers missing semantic cues to guide composed image retrieval, and the retrieved references are used by reference-based completion models.
To facilitate this task, we introduce a dataset containing 54K human-centric samples with associated album images. 
Experiments across multiple baselines demonstrate the difficulty of personalized completion and highlight the importance of identity-consistent reference retrieval.
\end{abstract}    
\section{Introduction}
\label{sec:intro}

With the rapid growth of personal photography, individuals now maintain extensive collections of personal images that capture their appearance, clothing, and daily environments. 
These personal albums~\cite{joon2015person,thomee2016yfcc100m} inherently contain rich identity-consistent cues, including facial structure, hairstyle, clothing style, and habitual backgrounds. 
Such collections provide valuable visual priors that can support personalized image editing~\cite{gu2024swapanything} and completion~\cite{Tsai2025completeme}. 
In particular, when restoring occluded regions in personal photographs, album images often contain complementary visual evidence that reflects the same identity across different scenes and viewpoints.

Unlike generic web images, personal albums exhibit strong identity consistency across images. 
A single individual may appear repeatedly across many photographs with similar facial characteristics, clothing preferences, and social contexts. 
These recurring visual patterns provide valuable cues that can support identity-aware image editing and restoration. 
However, effectively exploiting such cues remains challenging, as the most relevant reference image may be located anywhere within a large personal album and must be automatically identified from many candidate images.

When editing or completing a personal photo, users typically provide a mask indicating the region to be filled. 
Conventional inpainting and diffusion-based approaches~\cite{ju2024brushnet,suvorov2022resolution,lugmayr2022repaint} rely primarily on local context or category-level priors, which may generate visually plausible textures but often fail to preserve identity-specific details. 
Reference-based completion methods~\cite{zhou2021transfill,yang2023paint,chen2024zero,Tsai2025completeme} offer a promising alternative by conditioning the generation process on an additional reference image that provides visual evidence of the missing content. 
However, these approaches typically assume that an appropriate reference image is explicitly provided by the user. 
In real-world personal photo collections, such references are rarely available beforehand and must instead be discovered from a large album containing numerous candidate images.

Selecting a suitable reference image is itself a challenging problem. 
The retrieved reference must not only match the semantic content of the missing region but also preserve the identity and appearance of the same individual. 
In large personal albums containing hundreds or thousands of images, many candidates may share similar visual contexts while differing in identity, pose, or clothing. 
As a result, naive similarity-based retrieval often fails to identify the most appropriate reference, motivating the need for reasoning-guided retrieval strategies.

\begin{figure}[tp]
    \centering
    \includegraphics[width=\linewidth]{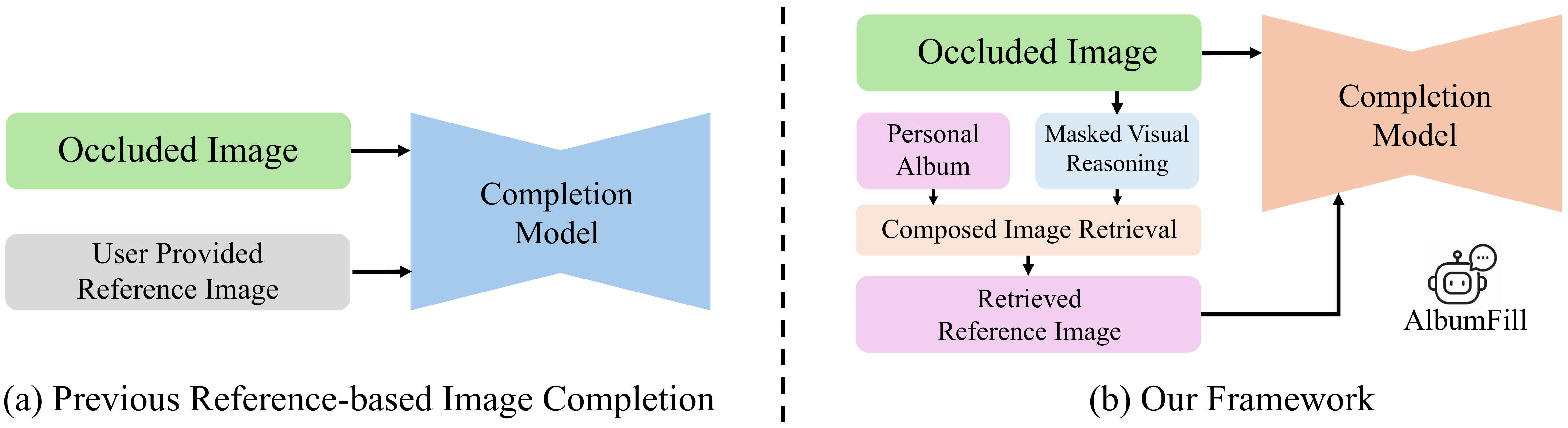}
    \caption{
\textbf{Comparison between previous reference-based image completion and our framework.}
(a) Previous methods assume that a suitable reference image is provided by the user, limiting their applicability when references are unavailable.
(b) Our framework automatically retrieves identity-consistent references from a personal photo album.
Given a masked input image, a reasoning module first infers the missing semantic cues, which are used to perform composed image retrieval from the album.
The retrieved reference image is then used by a completion model to synthesize the missing region.
}
    \label{fig:teaser}
    \vspace{-6mm}
\end{figure}

Vision-language models (VLMs)~\cite{bai2025qwen25,wang2024qwen2,wang2025internvl3,chen2024internvl,grattafiori2024llama,team2025gemma,zeng2025glm,li2023blip} provide a natural tool for addressing this challenge. 
By interpreting the visible context of the occluded image, VLMs can infer the likely semantics of the missing region and generate textual descriptions that capture the expected content. 
Such descriptions can serve as structured retrieval cues, enabling the system to search for identity-consistent reference images within the album that match both the inferred semantics and the individual's appearance.

Building upon this insight, we present \ourmethod, a framework for personalized image completion that performs reasoning-guided reference retrieval from personal albums. 
Given an occluded image and a personal album, our approach first employs a vision-language model to reason about the masked region and generate semantic cues describing the missing content. 
These cues are then used to perform composed image retrieval (CIR)~\cite{saito2023pic2word,jang2024spherical,lin2024fine,gu2024lincir,tang2024contexti2w,suo2024knowledge} over the album to identify reference images that are consistent with both the inferred semantics and the individual's identity. 
The retrieved reference image is subsequently combined with the masked input and passed to a reference-based completion model to synthesize the final result.

To support this task, we further introduce a new benchmark for personalized album-based completion. 
Our dataset contains 54,000 human-centric image samples paired with contextual captions and album references, enabling systematic evaluation of both reference retrieval and completion quality. 
Unlike conventional datasets that focus on generic reconstruction, our benchmark emphasizes identity-aware completion in personal photo collections.

Our work differs from prior prompt-based inpainting methods, which rely on generative priors to hallucinate missing content, as well as recent multimodal editing systems such as BAGEL~\cite{deng2025emerging} and Gemini 3 Pro~\cite{google2025geminiproimage} that generate edits directly from textual instructions. 
It also differs from existing reference-based completion approaches that assume a reference image is provided. 
Instead, we focus on the realistic setting where identity-consistent reference images must be automatically retrieved from personal albums before performing completion.

Our contributions are summarized as follows:
\begin{itemize}
    \item We propose \ourmethod, a framework that leverages vision-language reasoning to guide reference retrieval and enable identity-consistent image completion.
    \item We introduce a new benchmark for album-based personalized completion, containing 54K human-centric samples for evaluating retrieval and completion quality.
    \item We demonstrate through experiments that identity-consistent references are critical for personalized image completion, motivating the need for reference retrieval from personal photo collections.
\end{itemize}












\section{Related Work}
\label{sec:related}
\vspace{-2mm}
\noindent \textbf{Vision Language Model Reasoning.}
Recent advances in multimodal large language models (MLLMs)~\cite{bai2025qwen25,wang2024qwen2,wang2025internvl3,chen2024internvl,grattafiori2024llama,team2025gemma,zeng2025glm,li2023blip} highlight the role of instruction tuning~\cite{dai2023instructblip} and chain-of-thought (CoT)~\cite{wei2022chain} strategies in enabling robust visual reasoning. LLaVA~\cite{liu2023llava,liu2024llava15} aligns vision encoders with LLMs using curated multimodal instructions, yielding strong zero-shot generalization across tasks. Complementary CoT approaches extend reasoning supervision to the visual domain: Multimodal-CoT~\cite{zhang2024multimodal} separates rationale generation from answering, while Visual CoT~\cite{shao2024visual} provides large-scale data with region-level and step-by-step annotations to teach interpretable reasoning. Compositional CoT~\cite{mitra2024compositional} further leverages scene-graph–driven prompting to improve structured compositional inference. MM-REACT~\cite{yang2023mm} couples LLM reasoning with specialized vision tools for complex queries. Kosmos-2~\cite{peng2023kosmos} integrates explicit localization tokens to connect language to objects, strengthening referential reasoning. Finally, commercial systems~\cite{achiam2023gpt,comanici2025gemini,team2024gemini,lu2024deepseek} demonstrate scalable, real-time multimodal reasoning capabilities, underscoring the trend toward unified, instruction-following MLLMs.

\noindent \textbf{Zero-shot Composed Image Retrieval.}
Zero-shot Composed Image Retrieval (ZS-CIR)~\cite{saito2023pic2word,jang2024spherical,lin2024fine,gu2024lincir,tang2024contexti2w,suo2024knowledge} retrieves target images consistent with a reference image and user-provided textual modifications, without requiring extensive labeled triplets. ZS-CIR approaches leverage pre-trained vision-language models such as CLIP~\cite{radford2021learning} to combine visual and textual features within a shared embedding space. Recent textual inversion methods~\cite{baldrati2023zero,gu2024lincir} map reference image features into pseudo-word tokens or masked textual representations for query generation. Training-free frameworks~\cite{karthik2024visionbylanguage,sun2023ldre} further eliminate fine-tuning by using large language models to infer compositional queries directly. OSrCIR~\cite{tang2025osrcir} introduces a one-stage paradigm that employs multimodal large language models (MLLMs) for joint visual-textual reasoning, preserving full image context and achieving superior efficiency and intent alignment.

\begin{figure*}[tp]
    \centering
    \includegraphics[width=\linewidth]{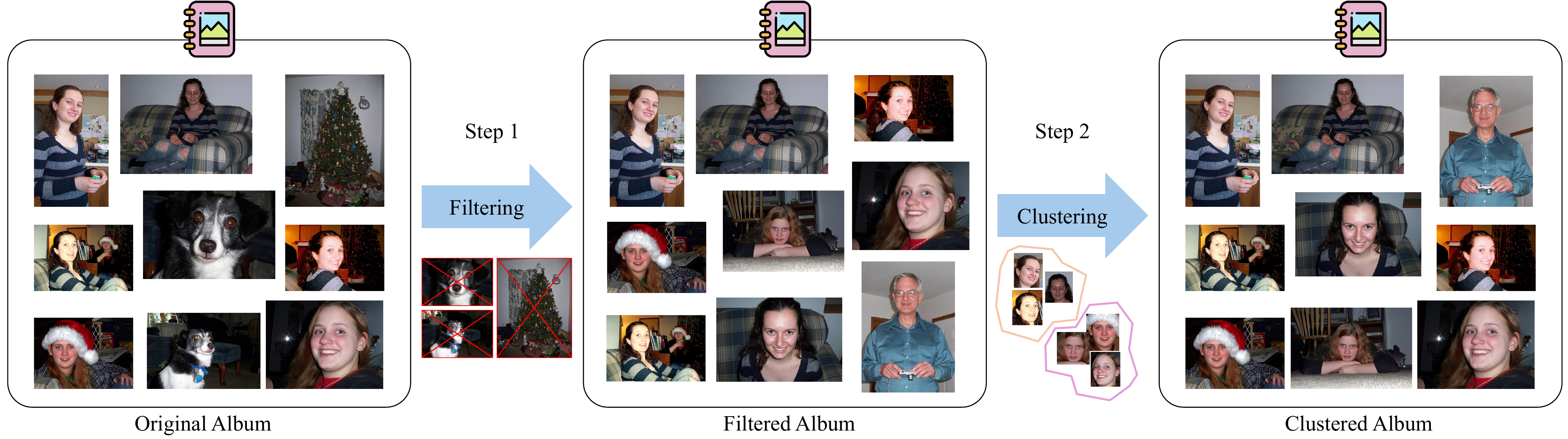}
    \caption{\textbf{Data generation pipeline for constructing our Album Dataset.} 
As describe in Sec.~\ref{sec:album_benchmark}, we start from the original CUFED album~\cite{wang2016event} (left), we first perform human detection and remove images without sufficiently large or clear human presence (Step~1: Filtering). 
We then apply identity clustering using InsightFace to group images by subject and select the dominant identity within each album (Step~2: Clustering). 
The resulting clustered album (right) contains only images of the target person, forming a coherent personal-level collection for identity-consistent retrieval and completion.
    }
    \label{fig:dataset_pipeline}
    \vspace{-6mm}
\end{figure*}

\noindent \textbf{Reference-based Inpainting.}
Reference-based image inpainting leverages external visual cues to produce semantically consistent completions. 
Early approaches focus on geometric alignment between source and target images. 
For example, TransFill~\cite{zhou2021transfill} aligns views through depth-aware multi-homography to preserve structural consistency. 
Recent methods increasingly adopt diffusion-based generation to incorporate reference information. 
Paint-by-Example~\cite{yang2023paint}, ObjectStitch~\cite{song2023objectstitch}, and IMPRINT~\cite{song2024imprint} integrate exemplar features through patch conditioning, content adaptors, or staged identity learning to better preserve appearance and semantics. 
Other approaches extend reference-guided editing to more flexible settings. 
AnyDoor~\cite{chen2024anydoor} enables zero-shot reference-guided synthesis, while LeftRefill~\cite{cao2024leftrefill} combines reference and target views as a unified input. 
MimicBrush~\cite{chen2024zero} performs localized reference editing using dual diffusion U-Nets trained in a self-supervised manner. 
CompleteMe~\cite{Tsai2025completeme} further improves identity preservation through a region-focused attention mechanism.
More recently, unified multimodal frameworks aim to support broader image editing tasks. 
UniReal~\cite{chen2025unireal} formulates image editing as discontinuous video generation to capture real-world dynamics. 
MIGE~\cite{tian2025mige} unifies subject-driven generation and instruction-based editing through a shared multimodal encoder, and EditMaster~\cite{zhang2025editmaster} combines textual and visual controls for multimodal editing. 
Large multimodal systems such as BAGEL~\cite{deng2025emerging} and Gemini 3 Pro~\cite{google2025geminiproimage} further support instruction-following image editing with visual references.

However, these approaches generally assume that appropriate reference images are explicitly provided. 
In realistic personal photo collections, references must instead be retrieved from large albums, requiring reasoning over incomplete visual inputs to identify identity-consistent references.
\section{AlbumFill Benchmark}
\label{sec:dataset}

\label{sec:album_benchmark}

To evaluate our framework in realistic personal-photo scenarios, we construct the \textbf{AlbumFill Benchmark}, a dataset designed to reflect the characteristics of real-world personal photo collections. 
We build our dataset upon CUFED~\cite{wang2016event}, which was originally designed for studying event-specific image importance and is constructed from the YFCC100M dataset~\cite{thomee2016yfcc100m}. 
CUFED contains 94,797 images organized into 1,883 folders, with each folder containing approximately 30 to 100 images. 
This album-like structure naturally reflects groups of photos captured within the same events or contexts, making it well suited for studying album-level visual relationships.

Since personal photo collections are typically human-centric and identity consistency plays a key role in personalized completion, we focus on images containing visible people and ensure that the masked regions correspond to identity-relevant areas.

\noindent \textbf{Step 1: Human Detection and Filtering.}  
We therefore detect people using YOLOv8~\cite{yolov8_ultralytics}. 
Bounding boxes smaller than 0.15 of the image height or width are discarded because they correspond to distant or insignificant figures. 
Images containing more than 20 people are removed to avoid overly crowded scenes.

\noindent \textbf{Step 2: Identity Clustering and Album Formation.}  
Next, we apply InsightFace~\cite{insightface2025} to cluster identities within each album. 
For each album, we select the dominant identity, defined as the person who appears in the largest number of images, as the target subject. 
All images that contain this identity are grouped to form one personal album that spans diverse poses, outfits, and backgrounds. 
Each image within an album can serve as a potential input image, while the remaining images act as candidate reference images for retrieval. 
Because multiple visually similar images of the same individual appear in each album under different poses, scenes, and viewpoints, identifying the most appropriate reference image becomes a non-trivial retrieval problem.

\noindent \textbf{Step 3: Mask Generation.}  
To construct completion tasks, we generate random occlusion masks on the selected input images. 
Masks are placed on human-centric regions to simulate realistic occlusions commonly observed in personal photographs, such as missing faces, bodies, or accessories. 
The masked image serves as the input to the completion model, while the original image is used as the ground-truth reference for evaluation.
After filtering and clustering, the final dataset contains \textbf{53,294 images grouped into 1,727 personal albums}. 
Each album corresponds to a single dominant identity with an average of \textbf{30.86 photos}. 
The dataset exhibits substantial diversity in pose, clothing, indoor and outdoor scenes, and background complexity, creating challenging conditions for reference-based completion and retrieval. 
Overall, the AlbumFill Benchmark provides a realistic testbed for evaluating reasoning-guided retrieval and identity-consistent image completion in personal photo collections. 
Additional dataset construction details are provided in the supplementary material.

\section{Method}
\label{sec:method}

\begin{figure}[tp]
    \centering
    \includegraphics[width=\linewidth]{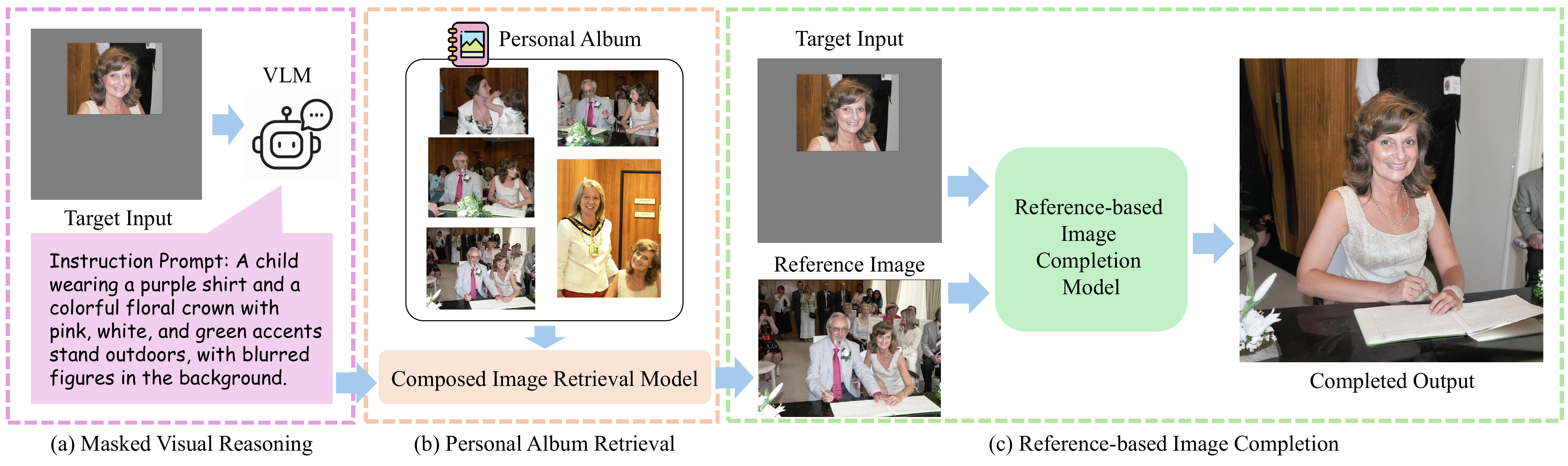}
    \caption{
\textbf{AlbumFill system overview.}
(a) Given a masked target image, a Vision-Language Model (VLM) performs \textit{masked visual reasoning} to generate a textual hypothesis describing the likely content behind the masked region. 
(b) The reasoning text and visible context are used to compose a multimodal query that retrieves the most semantically aligned and identity-consistent reference image from the user’s personal album. 
(c) A reference-based image completion model synthesizes the final output by integrating the masked target and the retrieved reference, producing an identity-faithful and contextually coherent restoration.
}
    \label{fig:system}
    \vspace{-6mm}
\end{figure}

\subsection{Overview}

We propose \ourmethod, a training-free framework that combines vision-language reasoning, album-based retrieval, and reference-guided image completion to enable personalized restoration from personal photo collections. 
Instead of fine-tuning or training new models, \ourmethod~connects pre-trained large-scale vision-language and generative components within a modular pipeline.
Given an input image $I_m$ with a missing region and a user's personal album $\mathcal{A} = \{I_1, I_2, ..., I_N\}$, the system operates in three sequential stages:
(1) \textbf{Masked visual reasoning}, where a vision-language model analyzes the incomplete image and infers the semantic content of the occluded region based on the visible context;
(2) \textbf{Album retrieval}, which searches the personal album for reference images that are consistent with the reasoning output and the visible identity cues; and
(3) \textbf{Reference-based completion}, which synthesizes the final image using both the masked input and the retrieved reference image.
This modular design allows flexible substitution of different vision-language models or completion models while requiring no additional task-specific training. 
The overall pipeline is illustrated in Fig.~\ref{fig:system}.

\subsection{Masked Visual Reasoning}

Existing vision-language models (VLMs) typically assume access to complete visual inputs when performing semantic reasoning. 
However, image editing and inpainting tasks involve intentionally occluded regions, requiring the model to infer plausible content behind the masked area based on the available visual context. 
To address this setting, we perform \textbf{masked visual reasoning}, where a VLM predicts the semantic content of the missing region using only the visible portion of the image together with its learned world knowledge.

Given a masked image $I_m$ and its binary mask $M$, the visible region is obtained as
\[
I_m^{vis} = I_m \odot (1 - M).
\]
We then prompt a pre-trained VLM with an instruction such as:
\textit{``Describe what likely exists in the masked region based on the visible context.''}
The VLM produces a reasoning hypothesis
\[
T_r = \text{VLM}(I_m^{vis}, \text{instruction}),
\]
where $T_r$ is a natural-language description containing semantic cues such as object category, pose, appearance, spatial relations, or activity. 
The full prompt templates used for reasoning are provided in the supplementary material.
This step reframes the pixel-level completion problem as a \textit{conceptual prediction task}, allowing the model to reason beyond directly observable content. 
Unlike conventional inpainting methods that extrapolate local textures, masked visual reasoning produces global structural priors. 
For example, the model may infer descriptions such as 
``the child’s missing hand holding a toy'' or 
``the upper part of the head wearing a floral crown.'' 
These semantic cues guide the subsequent album retrieval stage by providing structured hints about the missing content.

\subsection{Album Retrieval}
Given the reasoning hypothesis $T_r$, our goal is to retrieve album images that align with the inferred semantics while preserving the user's identity. 
Because each album in our benchmark contains images of the same dominant identity (Sec.~\ref{sec:dataset}), the retrieval process focuses on selecting references that best match the inferred content, pose, and scene context.

\noindent \textbf{Composed Query Embedding.}
To integrate visual context with reasoning cues, we construct a multimodal query embedding that combines the visible region of the masked image with the reasoning text:
\[
q = f_{\text{CLIP}}(I_m^{vis}, T_r),
\]
where $f_{\text{CLIP}}$ denotes a CLIP-style joint vision-language encoder. 
This composed embedding captures visual appearance cues from the visible image region (e.g., clothing color or hairstyle), contextual information such as indoor or outdoor environments, and semantic predictions produced during the reasoning stage. 
This formulation follows the standard composed image retrieval paradigm.

\noindent \textbf{Album-Level Retrieval.}
Each album image $I_i \in \mathcal{A}$ is encoded as
\[
v_i = f_{\text{CLIP}}(I_i).
\]
We compute the similarity between the query and each candidate image via cosine similarity:
\[
s_i = \cos(q, v_i).
\]
The top-$k$ most similar images are selected as
\[
\mathcal{R} = \text{Top-}k(\{s_i\}_{i=1}^{N}).
\]

In our default pipeline, the highest-ranked image $I_{\text{ref}}$ is used as the reference for the completion model. 
However, the full candidate set $\mathcal{R}$ can also be presented to the user as alternative reference options, allowing manual selection when multiple plausible references exist. 
This design provides both an automatic retrieval mechanism and an interactive reference selection interface.

By retrieving references from the personal album, this stage provides identity-consistent visual evidence for completion rather than relying solely on generative priors. 
The reasoning cues further narrow the retrieval space by describing the likely semantics of the occluded region, improving alignment between the retrieved reference and the missing content.

\subsection{Reference-based Completion}
Given the masked target image $I_m$ and the retrieved reference image $I_{ref}$, the final stage synthesizes the completed output using a reference-based diffusion inpainting model. 
We denote the completion module as $\mathcal{G}$, which produces the restored image as
\[
I_{out} = \mathcal{G}(I_m, I_{ref}).
\]

\noindent \textbf{Reference-Guided Conditioning.}
The inpainting model incorporates visual features from the reference image to guide the synthesis of the missing region. 
These reference cues provide fine-grained appearance information such as texture and color, structural information such as pose or silhouette, and scene-level attributes such as lighting and tone. 
Conditioning on both the masked input and the retrieved reference allows the model to generate content that is consistent with the identity and appearance observed in the album images.

\noindent \textbf{Consistency Preservation.}
To maintain spatial coherence, the completion model leverages both the visible context in $I_m$ and the appearance cues in $I_{ref}$. 
This combination encourages structural continuity along the mask boundary while preserving identity-specific characteristics within the synthesized region. 
As a result, the generated completion aligns with the surrounding scene context and the retrieved reference appearance.

Because the completion module is independent from the reasoning and retrieval stages, different reference-based inpainting models can be used within the same pipeline without additional task-specific training.

\section{Experiments}
\label{sec:experiment}
\subsection{Experimental Setups}
\noindent \textbf{Implementation Details.}
For masked visual reasoning, we evaluate three state-of-the-art Vision-Language Models (VLMs): Qwen3-VL-8B~\cite{yang2025qwen3}, InternVL3.5-8B~\cite{wang2025internvl3}, and LLaVA-1.6-7B~\cite{liu2024llava15}. 
All models are used in a zero-shot setting without any task-specific fine-tuning. 
We apply a unified instruction template that asks the model to infer the likely semantic content of the masked region based on the visible context of the image. 
Each masked input produces a single reasoning hypothesis, which is then used to guide the subsequent retrieval stage. 
Unless otherwise specified, we use Qwen3-VL-8B as the default reasoning model due to its stronger semantic grounding performance.

For the Composed Image Retrieval (CIR) module, we benchmark three representative zero-shot methods: iSEARLE~\cite{agnolucci2025isearle}, CIReVL~\cite{karthik2024visionbylanguage}, and LinCIR~\cite{gu2024lincir}. 
To ensure consistency across methods, all retrieval models use the same CLIP~\cite{radford2021learning} backbone, ViT-L/14~\cite{dosovitskiy2020image}, for encoding both composed queries and album images. 
Each album image is embedded and ranked using cosine similarity with respect to the composed query embedding. 
We retrieve the top-$k$ candidates ($k=5$ unless otherwise specified) and pass the highest-ranked reference image to the completion module. 
For reference-based completion, we evaluate three models: MimicBrush~\cite{chen2024zero}, CompleteMe~\cite{Tsai2025completeme}, and UniReal~\cite{chen2025unireal}. 
Each model is used with the default inference configurations provided by the official implementations. 


All experiments are conducted on machines equipped with NVIDIA A100 GPUs.
After benchmarking different components, our default pipeline uses Qwen3-VL-8B~\cite{yang2025qwen3} for masked visual reasoning, LinCIR~\cite{gu2024lincir} for composed image retrieval, and UniReal~\cite{chen2025unireal} for reference-guided completion. 
Unless otherwise specified, all reported results follow this configuration.

\noindent \textbf{Evaluation Metrics.}
We evaluate our framework along two key aspects of the AlbumFill pipeline: (1) the retrieval accuracy in the album retrieval stage, and (2) the visual fidelity and perceptual consistency of the completed images.

\noindent \textbf{Album Retrieval.}
For composed image retrieval (CIR), we follow standard evaluation protocols and report Recall@K and mean Average Precision at K (mAP@K) for $K \in \{1,5,10,25,50\}$. 
Recall@K measures whether relevant reference images appear within the top-$K$ retrieved results, while mAP@K evaluates the ranking quality among all relevant candidates within the album.

\noindent \textbf{Reference-based Completion.}
To assess the quality of generated completions, we measure both pixel-level fidelity and perceptual similarity. 
PSNR~\cite{hore2010image} and SSIM~\cite{wang2004image} evaluate low-level reconstruction accuracy between the generated output and the ground-truth image. 
To better capture perceptual and semantic consistency, we also report LPIPS~\cite{zhang2018unreasonable}, DINO similarity~\cite{caron2021emerging}, CLIP similarity~\cite{radford2021learning}, and DreamSim~\cite{fu2024dreamsim}.




\subsection{VLM-based Reasoning Evaluation}

To evaluate the quality of manipulation instructions produced during masked visual reasoning, we employ an automatic evaluation protocol using a strong Vision-Language Model (VLM) as an external evaluator. Specifically, we use Gemini 2.5 Pro~\cite{comanici2025gemini}.
The evaluator takes the masked image and the generated instruction as input and assesses whether the instruction provides useful cues for retrieving a correct reference image for completion. Unlike traditional caption evaluation, the goal is not linguistic similarity to ground truth but whether the instruction offers grounded and discriminative information for the retrieval stage.
To ensure independent assessment, the evaluation model is different from the reasoning models, avoiding potential bias from using the same model family for both generation and evaluation.

The evaluator scores each instruction along four dimensions: \textit{Evidence Grounding}, \textit{Structural Continuity}, \textit{Retrieval Discriminativeness}, and \textit{Instruction Format Quality}, each ranging from 0 to 20. These metrics measure whether the instruction is supported by visible evidence, specifies structural continuation across the mask, provides discriminative retrieval cues, and follows a concise manipulation-instruction format. The full evaluation prompt and rubric are provided in the supplementary material.

Table~\ref{tab:instruction_eval} reports the results of different reasoning models. Qwen3-VL-8B achieves the highest scores across all dimensions, producing more grounded and discriminative instructions that better support the downstream retrieval stage.

\begin{table}[t]
\centering
\small
\caption{\textbf{Evaluation of instruction quality using a VLM-based evaluator.}
Instruction quality is evaluated using Gemini~2.5~Pro~\cite{comanici2025gemini} as an external VLM evaluator to maintain independence from the reasoning models. We measure four aspects: 1) evidence grounding, 2) structural continuity, 3) retrieval discriminativeness, and 4) instruction format quality. Scores range from 0 to 20 for each dimension and are averaged across all evaluation samples. Qwen3-VL~\cite{yang2025qwen3} achieves the highest scores across all dimensions, indicating stronger grounding and more discriminative instructions, and is therefore selected as the reasoning module in our framework.}
\vspace{-2mm}
\centering
\resizebox{0.9\linewidth}{!} 
  {
\begin{tabular}{lcccc}
\toprule
Model & \shortstack{Evidence\\Grounding} & \shortstack{Structural\\Continuity} & \shortstack{Retrieval\\Discriminativeness} & \shortstack{Instruction Format\\Quality} \\
\midrule
LLaVA-1.6-7B~\cite{liu2024llava15}   & 11.00 & 11.53 &  8.90 & 18.05 \\
InternVL3.5-8B~\cite{wang2025internvl3} & 12.01 & 12.68 & 10.56 & 17.52 \\
Qwen3-VL-8B~\cite{yang2025qwen3}    & \textbf{15.23} & \textbf{14.88} & \textbf{13.75} & \textbf{18.66} \\
\bottomrule
\end{tabular}%
}
\label{tab:instruction_eval}
\vspace{-6mm}
\end{table}

\subsection{Quantitative Comparison on CIR}

We evaluate three composed image retrieval (CIR) methods, iSEARLE, CIReVL, and LinCIR, on a subset of 1,416 albums that contain sufficient valid query-reference pairs, while the full dataset contains 1,727 albums.
For each album, a masked query image is paired with a manipulation instruction generated by Qwen3-VL, which describes the expected content behind the masked region. 
This instruction serves as the language component for all CIR baselines. All methods use CLIP ViT-L/14 encoder for fair comparison.

Table~\ref{tab:cir_evaluation} reports Recall@k and mAP@k for $k \in \{1,5,10,25,50\}$. 
LinCIR achieves the best performance across all evaluation settings. 
At low retrieval depths ($k=1, 5, 10$), LinCIR obtains the highest Recall@k and mAP@k values. 
In particular, Recall@1 increases from 6.44 (iSEARLE) and 6.88 (CIReVL) to \textbf{7.00}. 
This result suggests that LinCIR forms more discriminative composed representations for masked-query retrieval.

At larger retrieval depths, the three methods show similar recall trends, reaching approximately 80 percent Recall@25. 
However, LinCIR consistently achieves the highest ranking quality, obtaining the best mAP scores across all evaluated depths, including \textbf{50.21} at $k=25$ and \textbf{57.69} at $k=50$. 
These results indicate that LinCIR maintains more stable ranking performance when a larger number of candidate references are considered.
iSEARLE and CIReVL exhibit comparable performance across most metrics, although both methods remain slightly below LinCIR in terms of top-$k$ retrieval accuracy and ranking quality. 
Overall, LinCIR provides the most reliable retrieval performance for masked-query CIR on the AlbumFill benchmark.

\begin{table*}[tp]
\small
\centering
  \caption{
\textbf{Quantitative comparison of composed image retrieval (CIR) performance on our personal album benchmark.}
We evaluate three representative CIR models: iSEARLE~\cite{agnolucci2025isearle}, CIReVL~\cite{karthik2024visionbylanguage}, and LinCIR~\cite{gu2024lincir}, under the masked-query setting, using Recall@K and mAP@K for $K=\{1,5,10,25,50\}$. 
Overall, LinCIR achieves the highest retrieval accuracy across all metrics, particularly at low $K$, indicating stronger fine-grained ranking ability. 
}
  \vspace{-2mm}
  \centering
  \label{tab:cir_evaluation}
    \resizebox{0.8\linewidth}{!} 
  {
  \centering
  \scriptsize
  \begin{tabular}{lcccccccc}
    \toprule
    Metric &  \multicolumn{4}{c}{Recall@k}  &  \multicolumn{4}{c}{mAP@k}\\
    \midrule
    Method & k=1 & k=5 & k=10 & k=25 & k=5 &k=10 & k=25 & k=50\\
    \midrule
    iSEARLE~\cite{agnolucci2025isearle}         & 6.44 & 28.42 & 46.70 & 80.59 & 21.61 & 31.73 & 48.55 & 56.19\\
    CIReVL~\cite{karthik2024visionbylanguage}   & 6.88 & 29.17 & 47.70 & 80.95 & 22.50 & 32.69 & 49.69 & 57.17\\
    LinCIR~\cite{gu2024lincir}                  & \textbf{7.00} & \textbf{29.63} & \textbf{48.13} & \textbf{81.21}  & \textbf{22.83} & \textbf{33.31} & \textbf{50.21} & \textbf{57.69}\\
  \bottomrule
  \end{tabular}%
  }
\vspace{-2mm}
\end{table*}
\begin{table*}[tp]
\small
\centering
  \caption{
Quantitative comparison on the AlbumFill benchmark (Sec.~\ref{sec:dataset}).
Methods are grouped into three categories: inpainting methods without reference images (top), MLLM-based methods (middle), and reference-based image completion methods (bottom).
Our framework adopts UniReal~\cite{chen2025unireal} as the completion backbone.
}
  \vspace{-2mm}
  \centering
  \label{tab:completion_quantitative}
    \resizebox{0.9\linewidth}{!} 
  {
  \centering
  \scriptsize
  \begin{tabular}{lcccccc}
    \toprule
    Method & CLIP $\uparrow$ & DINO $\uparrow$ & DreamSim $\downarrow$ & LPIPS $\downarrow$ & PSNR $\uparrow$ & SSIM $\uparrow$  \\
    \midrule
    BrushNet~\cite{ju2024brushnet} w/o Prompt & 87.29 & 87.82 & 0.162 & 0.233 & 18.35 & 0.654 \\
    BrushNet~\cite{ju2024brushnet} with Prompt & 88.38 & 88.19 & 0.138 & 0.229 & 18.81 & 0.655\\ 
    \midrule
    BAGEL~\cite{deng2025emerging} & 77.52 & 66.29 & 0.364 & 0.652 & 9.47 & 0.174\\
    Gemini 3 Pro~\cite{google2025geminiproimage} & 89.35 & 88.26 & 0.147 & 0.322 & 15.83 & 0.511\\
    \midrule
    MimicBrush~\cite{chen2024zero} & 91.24 & 90.55 & 0.115 & 0.115 & 20.11 & \textbf{0.821} \\
    CompleteMe~\cite{Tsai2025completeme} & 90.70 & 91.70 & 0.102 & 0.212 & 20.81 & 0.768 \\
    UniReal~\cite{chen2025unireal} &\textbf{94.70}&\textbf{95.71}&\textbf{0.055}&\textbf{0.142}&\textbf{22.60}&0.790\\
  \bottomrule
  \end{tabular}%
  }
\vspace{-6mm}
\end{table*}
\begin{figure}[tp]
    \centering
    \includegraphics[width=0.95\linewidth]{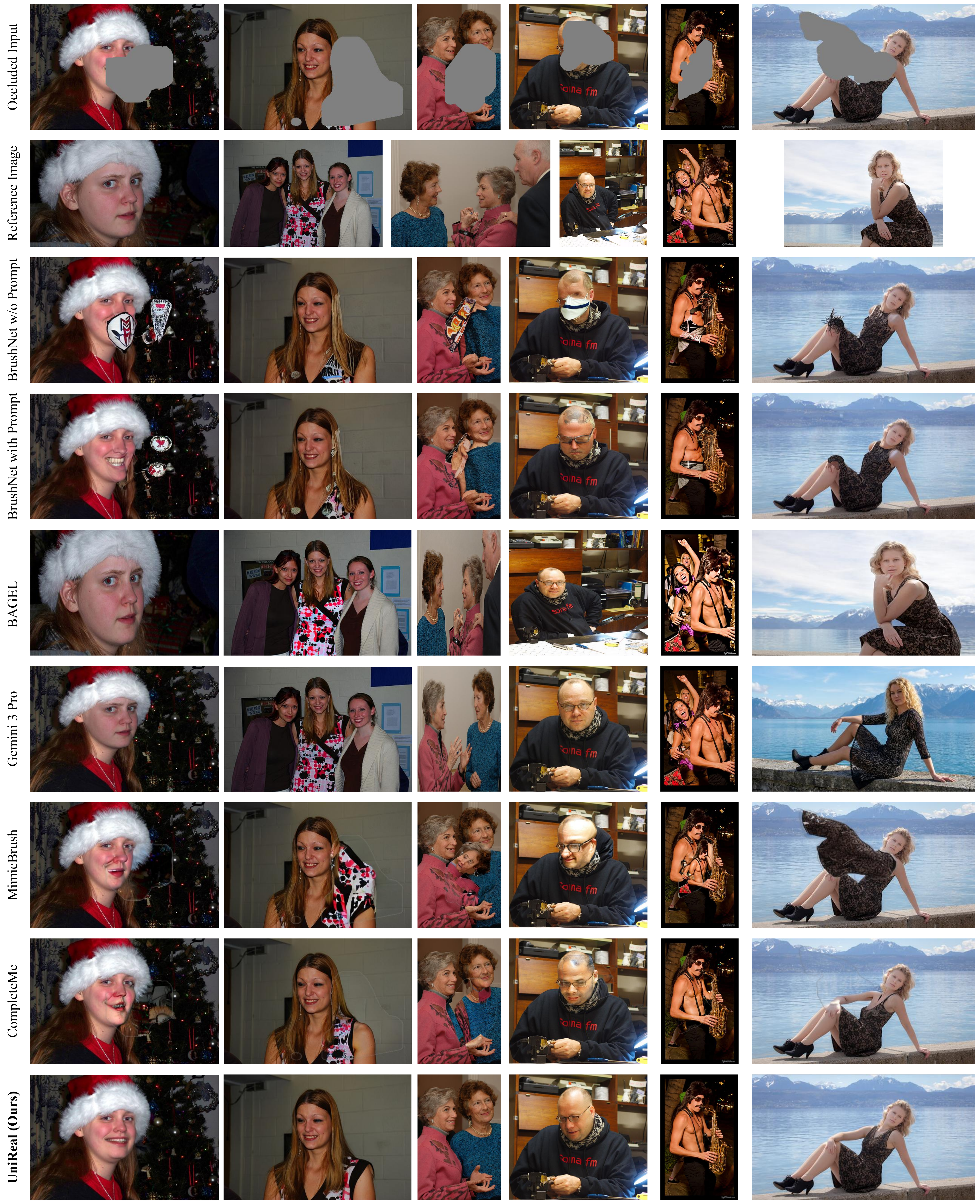}
    \vspace{-2mm}
    \caption{
    \textbf{Visual comparison with different categories of completion methods on our Album Benchmark~(Sec.~\ref{sec:album_benchmark}).}
We compare our method with three types of baselines: prompt-based inpainting (BrushNet~\cite{ju2024brushnet} with and without prompts), MLLM-based image editing models (BAGEL~\cite{deng2025emerging} and Gemini 3 Pro~\cite{google2025geminiproimage}), and reference-based completion methods (MimicBrush~\cite{chen2024zero} and CompleteMe~\cite{Tsai2025completeme}).
Prompt-based methods often produce plausible textures but fail to preserve identity.
MLLM-based approaches struggle with identity consistency or structural continuity under large occlusions.
Reference-based methods rely heavily on the quality of the provided reference image.
In contrast, our method retrieves identity-consistent references from personal albums and produces more coherent completions. For more qualitative comparisons, please refer to our supplementary material.
    }
    \label{fig:qualitative}
\end{figure}

\subsection{Quantitative and Qualitative Comparison on Image Completion}

We evaluate representative image completion approaches on the AlbumFill benchmark, including reference-based inpainting models and prompt-driven generative systems. 
Table~\ref{tab:completion_quantitative} reports CLIP similarity, DINO similarity, DreamSim distance, LPIPS, PSNR, and SSIM. 
All methods use the same masked inputs and reference images for fair comparison, as shown in Fig.~\ref{fig:qualitative}.

Among the evaluated methods, UniReal~\cite{chen2025unireal} achieves the best overall performance, obtaining the highest CLIP (94.70) and DINO (95.71) scores, the lowest DreamSim distance (0.055), and the best PSNR (22.60). 
MimicBrush~\cite{chen2024zero} produces reasonable structural alignment but lower perceptual consistency, while CompleteMe~\cite{Tsai2025completeme} shows comparable reconstruction fidelity.

We also evaluate prompt-driven baselines to examine whether large multimodal generative models can directly solve masked completion without specialized reference-guided architectures. 
For BrushNet~\cite{ju2024brushnet}, we test both a default setting and a prompt generated by the masked reasoning stage using Qwen3-VL. 
Although prompt guidance slightly improves results, the gap with reference-based inpainting models remains large.

We further evaluate BAGEL~\cite{deng2025emerging} and Gemini~3~Pro~\cite{google2025geminiproimage} with the same reference images and detailed instructions. 
Despite the prompts, these models often produce unstable outputs, such as incorrect aspect ratios, returning the reference image, or directly combining the masked and reference images (see Fig.~\ref{fig:qualitative}), leading to significantly worse scores. The full prompts used for these models are provided in the supplementary material.

Overall, reference-based inpainting models remain substantially more reliable for personalized image completion. 
Among them, UniReal provides the most consistent performance across perceptual and reconstruction metrics, and is therefore adopted as the completion module in our framework.

\begin{table*}[tp]
\small
\centering
  \caption{
\textbf{Effect of reasoning prompts on composed image retrieval under different mask ratios.}
We evaluate LinCIR with and without reasoning-generated prompts produced by Qwen3-VL. 
Mask ratios are categorized as small (<20\%), medium (20–50\%), and large (>50\%). 
Incorporating reasoning prompts consistently improves Recall@k and mAP@k across all mask regimes, with larger gains observed when the visible region becomes more limited.}
  \vspace{-2mm}
  \centering
  \label{tab:reasoning_ablation}
    \resizebox{0.8\linewidth}{!} 
  {
  \centering
  \scriptsize
  \begin{tabular}{lcccccccc}
    \toprule
    Metric &Reasoning Prompt& Mask Ratio & \multicolumn{3}{c}{Recall@k}  &  \multicolumn{3}{c}{mAP@k}\\
    \midrule
    Method & & & k=5 & k=10 & k=25 & k=5 &k=10 & k=25\\
    \midrule
    LinCIR~\cite{gu2024lincir} &    &  small  & 30.70 & 48.90 & 81.46 & \textbf{24.18} & 34.31 & 50.77 \\
    LinCIR~\cite{gu2024lincir} & \checkmark & small & \textbf{31.08} & \textbf{49.74} & \textbf{82.28} & 24.02 & \textbf{34.66} & \textbf{51.41} \\
    \midrule
    LinCIR~\cite{gu2024lincir}     &        &   medium  & 25.91 & 43.18 & 77.28 & 19.82 & 29.32 & 46.31\\
    LinCIR~\cite{gu2024lincir}    &     \checkmark    &  medium   & \textbf{26.80} & \textbf{45.10} & \textbf{78.65} & \textbf{20.62} & \textbf{30.93} & \textbf{48.29}\\
    \midrule
    LinCIR~\cite{gu2024lincir}     &        &   large  & 20.66 & 36.99 & 79.00 & 14.85 & 22.00 & 38.36\\
    LinCIR~\cite{gu2024lincir}      &   \checkmark    &   large  & \textbf{22.47} & \textbf{39.47} & \textbf{79.63} & \textbf{16.22} & \textbf{24.56} & \textbf{40.96}\\
  \bottomrule
  \end{tabular}%
  }
\vspace{-6mm}
\end{table*}

\subsection{Ablation Studies}
\vspace{-2mm}
\noindent \textbf{Effectiveness of the Reasoning Stage.}
We conduct an ablation study to evaluate the impact of the masked visual reasoning stage on composed image retrieval. 
To isolate its effect, we fix the retrieval model to LinCIR and use Qwen3-VL as the reasoning model, comparing retrieval performance with and without the reasoning-generated prompt.
To analyze the effect of occlusion severity, we divide the benchmark by mask ratio: \textit{small} (<20\%), \textit{medium} (20–50\%), and \textit{large} (>50\%).

Table~\ref{tab:reasoning_ablation} reports Recall@k and mAP@k under these settings. 
Across all mask ratios, incorporating the reasoning prompt consistently improves retrieval performance. 
For example, Recall@5 increases from 30.70 to \textbf{31.08} for small masks, from 25.91 to \textbf{26.80} for medium masks, and from 20.66 to \textbf{22.47} for large masks.
These results indicate that reasoning-derived instructions provide additional semantic cues that help the CIR model better align masked inputs with relevant album images, with larger gains observed as occlusion increases.

\begin{table*}[tp]
\small
\centering
  \caption{
\textbf{Wrong-reference ablation under different mask ratios.}
The retrieval stage remains unchanged (Qwen3-VL + LinCIR retrieves from the correct album), while the completion reference is replaced with a random image from a different album (wrong identity). 
Mask ratios are categorized as small (<20\%), medium (20–50\%), and large (>50\%). 
Using a wrong-identity reference consistently degrades completion quality, with the largest performance drop observed under heavy occlusion.
}
  \vspace{-2mm}
  \centering
  \label{tab:wrong_reference_ablation}
    \resizebox{0.85\linewidth}{!} 
  {
  \centering
  \scriptsize
  \begin{tabular}{lcccccccc}
    \toprule
    Metric & Wrong Reference & Mask Ratio & CLIP $\uparrow$ & DINO $\uparrow$ & DreamSim $\downarrow$ & LPIPS $\downarrow$ & PSNR $\uparrow$ & SSIM $\uparrow$\\
    \midrule
    UniReal~\cite{chen2025unireal}  & \checkmark  &  small  & 96.01 & 97.30 & 0.037 & 0.114 & 24.00 & 0.827 \\
    UniReal~\cite{chen2025unireal} & & small & \textbf{96.65} & \textbf{97.72} & \textbf{0.030} & \textbf{0.110} & \textbf{24.32} & \textbf{0.829} \\
    \midrule
    UniReal~\cite{chen2025unireal}   &     \checkmark     &   medium  & 89.02 & 90.84 & 0.121 & 0.202 & 18.65 & 0.718\\
    UniReal~\cite{chen2025unireal} &       &  medium   & \textbf{91.75} & \textbf{93.09} & \textbf{0.092} & \textbf{0.193} & \textbf{19.24} & \textbf{0.724}\\
    \midrule
    UniReal~\cite{chen2025unireal}   &    \checkmark      &   large  & 69.89 & 66.99 & 0.348 & 0.389 & 14.10 & 0.508\\
    UniReal~\cite{chen2025unireal}    &      &   large  & \textbf{78.35} & \textbf{75.59} & \textbf{0.263} & \textbf{0.378} & \textbf{14.48} & \textbf{0.513}\\
  \bottomrule
  \end{tabular}%
  }
\vspace{-6mm}
\end{table*}

\noindent \textbf{Ablation on Identity-Consistent References.}

We perform a wrong-reference ablation to evaluate the importance of identity-consistent references in the completion stage. 
To isolate this effect, the retrieval stage remains unchanged: references are still retrieved from the correct album using the Qwen3-VL + LinCIR pipeline, but during completion the retrieved reference is replaced with a random image from a different album.
To analyze the impact of occlusion severity, we group samples by mask ratio: \textit{small} (<20\%), \textit{medium} (20–50\%), and \textit{large} (>50\%). 

Table~\ref{tab:wrong_reference_ablation} reports the results using UniReal as the completion model.
Across all mask ratios, replacing the reference with a wrong-identity image consistently degrades completion quality. 
For example, CLIP similarity drops from \textbf{96.65} to 96.01 for small masks, from \textbf{91.75} to 89.02 for medium masks, and from \textbf{78.35} to 69.89 for large masks.
These results show that identity-consistent references become increasingly important as the mask ratio grows, highlighting the role of personalized reference retrieval for reliable album-based image completion.

\section{Conclusion}
\label{sec:conclusion}

We present~\ourmethod, a reasoning-guided framework for personalized image completion that integrates masked-region reasoning, album-based retrieval, and reference-based completion. 
The framework decomposes the task into reasoning, retrieval, and completion, enabling the system to infer missing semantics, retrieve identity-consistent references from personal albums, and synthesize coherent results.
To support this task, we introduce the~\ourmethod~benchmark, which reflects the diversity and identity consistency of real-world personal photo collections. 
Experiments show that reasoning-derived instructions improve composed image retrieval under occlusion, while ablation studies demonstrate that identity-consistent references are crucial for successful completion, especially when large regions are missing.
Overall, our results highlight the importance of combining semantic reasoning with album-based retrieval for personalized image completion.



%
%
\bibliographystyle{splncs04}
\bibliography{main}

\end{document}